\def\year{2021}\relax
\documentclass[letterpaper]{article} 
\usepackage{aaai21}  
\usepackage{times}  
\usepackage{helvet} 
\usepackage{courier}  
\usepackage[hyphens]{url}  
\usepackage{graphicx} 
\def\UrlFont{\rm}  
\usepackage{natbib}  
\usepackage{caption} 
\usepackage{multirow}
\usepackage{comment}

\usepackage[table,xcdraw]{xcolor}
\title{COVID-19 Outbreak Prediction and Analysis using Self Reported Symptoms}
\author{Rohan Sukumaran\textsuperscript{$^*$1} \quad Parth Patwa\textsuperscript{$^*$1} \quad Sethuraman T V\textsuperscript{$^*$1} \quad Sheshank Shankar\textsuperscript{1} \\ \quad Rishank Kanaparti\textsuperscript{1} \quad Joseph Bae\textsuperscript{1,2} \quad Yash Mathur\textsuperscript{1} \quad Abhishek Singh\textsuperscript{1,4} \quad Ayush Chopra\textsuperscript{1,4} \quad Myungsun Kang\textsuperscript{1} \quad Priya Ramaswamy\textsuperscript{1,3} \quad Ramesh Raskar\textsuperscript{1,4} \\ \textsuperscript{1}PathCheck Foundation \quad \textsuperscript{2}Stony Brook Medicine \quad \\ \textsuperscript{3}University of California San Francisco \quad  \textsuperscript{4}MIT Media Lab \\

\{rohan.sukumaran, parth.patwa, sethu.ramantv\}@pathcheck.org 

}

\begin{document}

\maketitle
\renewcommand{\thefootnote}{\fnsymbol{footnote}}
\footnotetext[1]{Equal contribution.}
\renewcommand*{\thefootnote}{\arabic{footnote}}
\setcounter{footnote}{0}
\begin{abstract}
It is crucial for policy makers to understand the community prevalence of COVID-19 so combative resources can be effectively allocated and prioritized during the COVID-19 pandemic. Traditionally, community prevalence has been assessed through diagnostic and antibody testing data. However, despite the increasing availability of COVID-19 testing, the required level has not been met in most parts of the globe, introducing a need for an alternative method for communities to determine disease prevalence. This is further complicated by the observation that COVID-19 prevalence and spread varies across different spatial, temporal and demographics. In this study, we understand trends in the spread of COVID-19 by utilizing the results of self-reported COVID-19 symptoms surveys as an alternative to COVID-19 testing reports. This allows us to assess community disease prevalence, even in areas with low COVID-19 testing ability. Using individually reported symptom data from various populations, our method predicts the likely percentage of population that tested positive for COVID-19. We do so with a Mean Absolute Error (MAE) of 1.14 and Mean Relative error (MRE) of 60.40\% with 95\% confidence interval as (60.12, 60.67). This implies that our model predicts +/- 1140 cases than original in a population of 1 million. In addition, we forecast the location-wise percentage of the population testing positive for the next 30 days using self-reported symptoms data from previous days. The MAE for this method is as low as 0.15 (MRE of 23.61\%  with 95\% confidence interval as (23.6, 13.7)) for New York. We present analysis on these results, exposing various clinical attributes of interest across different demographics. Lastly, we qualitatively analyse how various policy enactments (testing, curfew) affect the prevalence of COVID-19 in a community. 

\end{abstract}

\section{Introduction}
The rapid progression of the COVID-19 pandemic has provoked large-scale data collection efforts on an international level to study the epidemiology of the virus and inform policies. Various studies have been undertaken to predict the spread, severity, and unique characteristics of the COVID-19 infection, across a broad range of clinical, imaging, and population-level datasets ~\cite{Gostic,Liang,Menni,Shi}. For instance, \cite{Menni} uses self-reported data from a mobile app to predict a positive COVID-19 test result based upon symptom presentation. Anosmia was shown to be the strongest predictor of disease presence, and a model for disease detection using symptoms-based predictors was indicated to have a sensitivity of about 65\%. Studies like \cite{Parma} have shown that ageusia and anosmia are widespread sequelae of COVID-19 pathogenesis. From the onset of COVID-19 there also has been significant amount of work in mathematical modeling to understand the outbreak under different situations for different demographics \cite{menni2020real, saad2020immune, wilder2020tracking}. Although these works primarily focus on population level the estimation of different transition probabilities to move between compartments is challenging.  

Carnegie Mellon University (CMU) and the University of Maryland (UMD) have built chronologically aggregated datasets of self-reported COVID-19 symptoms by conducting surveys at national and international levels \cite{UMDDataset,CMUDataset}. The surveys contain questions regarding whether the respondent has experienced several of the common symptoms of COVID-19 (e.g. anosmia, ageusia, cough, etc.) in addition to various behavioral questions concerning the number of trips a respondent has taken outdoors and whether they have received a COVID-19 test.

In this work, we perform several studies using the CMU, UMD and OxCGRT \cite{ UMDDataset, CMUDataset, OxfordDataset} datasets. Our experiments examine correlations among variables in the CMU data to determine which symptoms and behaviors are most correlated to high percentage of Covid Like Illness (CLI). We see how the different symptoms impact the percentage of population with CLI across different spatio-temporal and demographic (age, gender) settings. We also predict the percentage of population who got tested positive for COVID-19 and achieve 60\% Mean Relative Error.  
 Further, our experiments involve time-series analysis of these datasets to forecast CLI over time. Here, we identify how different spatial window trends vary across different temporal windows. We aim to use the findings from this method to understand the possibilities of modelling CLI for geographic areas in which data collection is sparse or non-existent. Furthermore, results from our experiments can potentially guide public health policies for COVID-19. Understanding how the disease is progressing can help the policymakers to introduce non pharmaceutical interventions (NPIs) and also help them understand how to distribute critical resources (medicines, doctors, healthcare workers, transportation and more). This could now be done based on the insights provided by our models, instead of relying completely on clinical testing data. Prediction of outbreak using self reported symptoms can also help reduce the load on testing resources. \newline
 
Using self reported symptoms collected across spatio-temporal windows 
to understand the prevalence and outbreak of COVID-19 is the first of its kind to the best of our knowledge.


\section{Datasets}
The \textbf{CMU Symptom Survey} aggregates the results of a survey run by CMU \cite{CMUDataset} which was distributed across the US to {\textasciitilde{}}70k random Facebook users daily.
COVIDcast gathers data from the survey and dozens of sources and produces a set of indicators which can inform our reasoning about the pandemic. Indicators are produced from these raw data by extracting a metric of interest, tracking revisions, and applying additional processing like reducing noise, adjusting temporal focus, or enabling more direct comparisons.

A few of which are

- 7 Public’s Behavior Indicators like People Wearing Masks and At Away Location 6hr+

- 3 Early Indicators like COVID-Related Doctor Visits and COVID-Like Symptoms in Community

- 4 Late Indicators like COVID Cases, COVID Deaths, COVID Antigen Test Positivity (Quidel) and Claims-Based COVID Hospital Admissions

 It has 104 columns, including weighted (adjusted for sampling bias), unweighted signals, demographic columns (age, gender etc) for county and state level data. We use the data from Apr. 4, '20 to Sep. 11, '20. This data is henceforth referred to as the CMU dataset in the paper.  \newline

The \textbf{UMD Global Symptom Survey} aggregates the results of a survey conducted by the UMD through Facebook \cite{UMDDataset}.The survey is available in 56 languages. A representative sample of Facebook users is invited on a daily basis to report on topics including, for example, symptoms, social distancing behavior, vaccine acceptance, mental health issues, and financial constraints. Facebook provides weights to reduce nonresponse and coverage bias. Country and region-level statistics are published daily via public API and dashboards, and microdata is available for researchers via data use agreements. Over half a million responses are collected daily.
We use the data of 968 regions, available from May 01 to September 11. There are 28 unweighted signals provided, as well as a weighted form (adjusted for sampling bias). These signals include self reported symptoms, exposure information, general hygiene etc.  \newline


The \textbf{Oxford COVID-19 Government Response Tracker (OxCGRT)} \cite{OxfordDataset} contains government COVID-19 policy data as a numerical scale value representing the extent of government action. OxCGRT collects publicly available information on 20 indicators of government response. This information is collected by a team of over 200 volunteers from the Oxford community and is updated continuously. Further, they also include statistics on the number of reported Covid-19 cases and deaths in each country. These are taken from the JHU CSSE data repository for all countries and the US States.

Here, for the timeseries and one-on-one predictions, we make use of 80\% of the entire data for training and use the remaining set aside 20\% for the testing purpose. The 80-20 split is random. 

Similar self reported data and survey data has been used by \cite{rodriguez2020steering, Rodriguezdeep, garcia2021estimating} for understanding the pandemic and drawing actionable insights.\\




The \textbf{Prevalence of Self-Reported Obesity by State and Territory, BRFSS, 2019- CDC} \cite{CDC} is a dataset published by CDC containing the aggregated self reported obesity values. The values are present at a granularity of state level and contains 3 columns corresponding to the name of the State , Obesity values and Confidence intervals (95\%). This dataset contains other details like Race, Ethnicity , Food habits etc which can used for further analysis.

\section{Method and Experiments}

\textbf{Correlation Studies}: Correlation between features of the dataset provides crucial information about the features and the degree of influence they have over the target value. We conduct correlation studies on different sub groups like symptomatic, asymptomatic and varying demographic regions in the CMU dataset to the discover relationships among the signals and with the target variable. We also investigate the significance of obesity and population density on the susceptibility to COVID-19 at state level~\cite{CDC}. Please refer to the Appendix for more information.  \newline




\textbf{Feature Pruning}: We drop demographic columns such as date, gender, age etc. Next we drop the unweighted columns because their weighted counterparts exist. We also drop features like percentage of people who got tested negative, weighted percentage of people who got tested positive etc as these are directly related to testing and would make the prediction trivial. Further, we drop derived features like estimated percentage of people with influenza-like illness because they were not directly reported by the respondents. Finally, we drop some features which calculate mean (such as average number of people in respondent’s household who have Covid Like Illness) because their range was in the order of $10^{50}$. After the entire process we are left with 36 features. The selected feature list is provided in Appendix. \newline

\textbf{Outbreak Prediction}: We predict the percentage of the population that tested positive (at a state level) from the CMU dataset. After feature pruning as mentioned above, we are left with 36 input signals. We rank these 36 signals according to their \textit{f\_regression} \cite{Freg} (\textit{f\_statistic} of the correlation to the target variable) and predict the target variable using the top \textit{n} ranked features. We experiment with \textit{top n features} value from 1 to 36 for various demographic groups. We train Linear Regression \cite{linear_regression}, Decision Tree \cite{dtrees} and Gradient Boosting \cite{gdbt} models. All the models are implemented using scikit-learn \cite{scikit-learn}.  \newline

\textbf{Time Series Analysis}: We predict the percentage of people that tested positive using the CMU dataset and percentage of people with CLI with the UMD dataset. Here, we make use of the top 11 features (according to their ranking obtained in outbreak prediction) from the CMU (36) and UMD (56) datasets for multivariate multi-step time series forecasting. Given the data is spread across different spatial windows (geographies) at a state level, we employ an agglomerative clustering method independently on symptoms and behavioural/external patterns, and sample locations which are not in the same cluster for our analysis. Using the Augmented Dickey-Fuller test \cite{cheung1995lag} we found the time series samples for these spatial windows to be stationary. Furthermore, we bucket the data based on the age and gender of the respondents, to provide granular insights on the model performance on various demographics. With a total of 12 demographic buckets [(age, gender) pairs] available, we use a Vector Auto Regressive (VAR) \cite{holden1995vector} model and an LSTM \cite{gers1999learning} model for the experiments. Furthermore, we qualitatively look at the impact of government policies (contact tracing, etc) on the spread of the virus.

\section{Results and Discussion}
\label{sec:results}
\begin{figure}[t!]
    \includegraphics[scale=0.46]{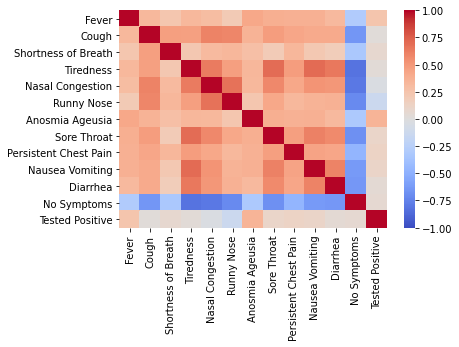}
    \caption{Correlation amongst self reported symptoms and \% tested COVID positive.}
    \label{fig:cmucorr}
\end{figure}

\begin{figure}[t!]
    \includegraphics[width=0.7\linewidth]{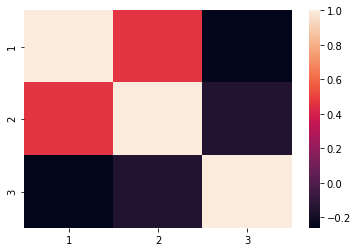}
  \caption{Correlation between the people having contact with someone having CLI and People tested positive. Here the attribute (1) = \% of people who had contact with someone having COVID-19, (2) = \% of people tested positive, (3) = \% of people who avoided contact all/most of the time  }
    \label{fig:covid_contact}
\end{figure}

\textbf{Correlation Studies:} State level analysis revealed a mild positive correlation, having an R value of 0.24 and a P value of the order of -257, between the percentage of people tested positive and statewide obesity level. The P  value was  Here the obesity is defined as BMI$>$ $30.0$ \cite{BMI}.These results are consistent with prior clinical studies like \cite{Chan2020} and indicate that further research required to see if lack of certain nutrients like Vitamin B, Zinc, Iron or having a BMI$>$ 30.0 could make an individual more susceptible to COVID-19. Figure \ref{fig:cmucorr} shows the correlation amongst multiple self reported symptoms and the symptoms having a significant positive correlations are highlighted. This clearly reveals that Anosmia, Ageusia  and fever are relatively strong indicators of COVID-19. From Figure \ref{fig:covid_contact}, we see that contact with a COVID-19 positive individual is strongly correlated with testing COVID-19 positive. Conversely, the percentage of population who avoid outside contact and the percentage of population testing positive for COVID-19 have a negative correlation. We also find a mild positive correlation between population density and percentage of population reporting COVID-19 positivity, which indicate easier transmission of the virus in congested environment. These observations reaffirm the highly contagious nature of the virus and the need for social distancing.  \newline 

The above results motivate us to estimate the \% of people tested COVID-19 positive based on \% of people who had a direct contact with anyone who recently tested positive. In doing so, we achieve an Mean Relative Error (MRE) of 2.33\% and Mean Absolute Error (MAE) of 0.03.  \newline

\begin{table}[]
    \centering
\begin{tabular}{|l|l|l|l|l|}
\hline
\textbf{Demographic} & \textbf{best n} & \textbf{MAE} & \textbf{MRE} & \textbf{CI} \\ \hline
Entire           & 35              & 1.14         & 60.40  & (60.12, 60.67)      \\ \hline
Male          & 34              & 1.38         & 78.14  &  (77.67, 78.62)
        \\ \hline 
Female        & 36              & 1.10         &  56.89 & (56.48, 57.30)        \\ \hline
Age 18-34      & 30              & 1.23         & 66.35 & (65.59, 67.12)         \\ \hline
Age 35-54     & 35              & 1.29         & 67.59  & (67.13, 68.04)        \\ \hline
Age 55+      & 33              & 1.20        & 66.40   & (65.86, 66.94)      \\ \hline
\end{tabular}
    \caption{Results of gradient boosting model for prediction of \% of population tested positive across demographics. The 95\% confidence interval (CI) for Mean Relative Error (MRE) is calculated on 20 runs (data shuffled randomly every time). the MRE and Mean Absolute Error (MAE) are average of 20 runs.}
    \label{table:symptoms}
\end{table}

\begin{figure}[t!]
    \includegraphics[width=0.8\linewidth]{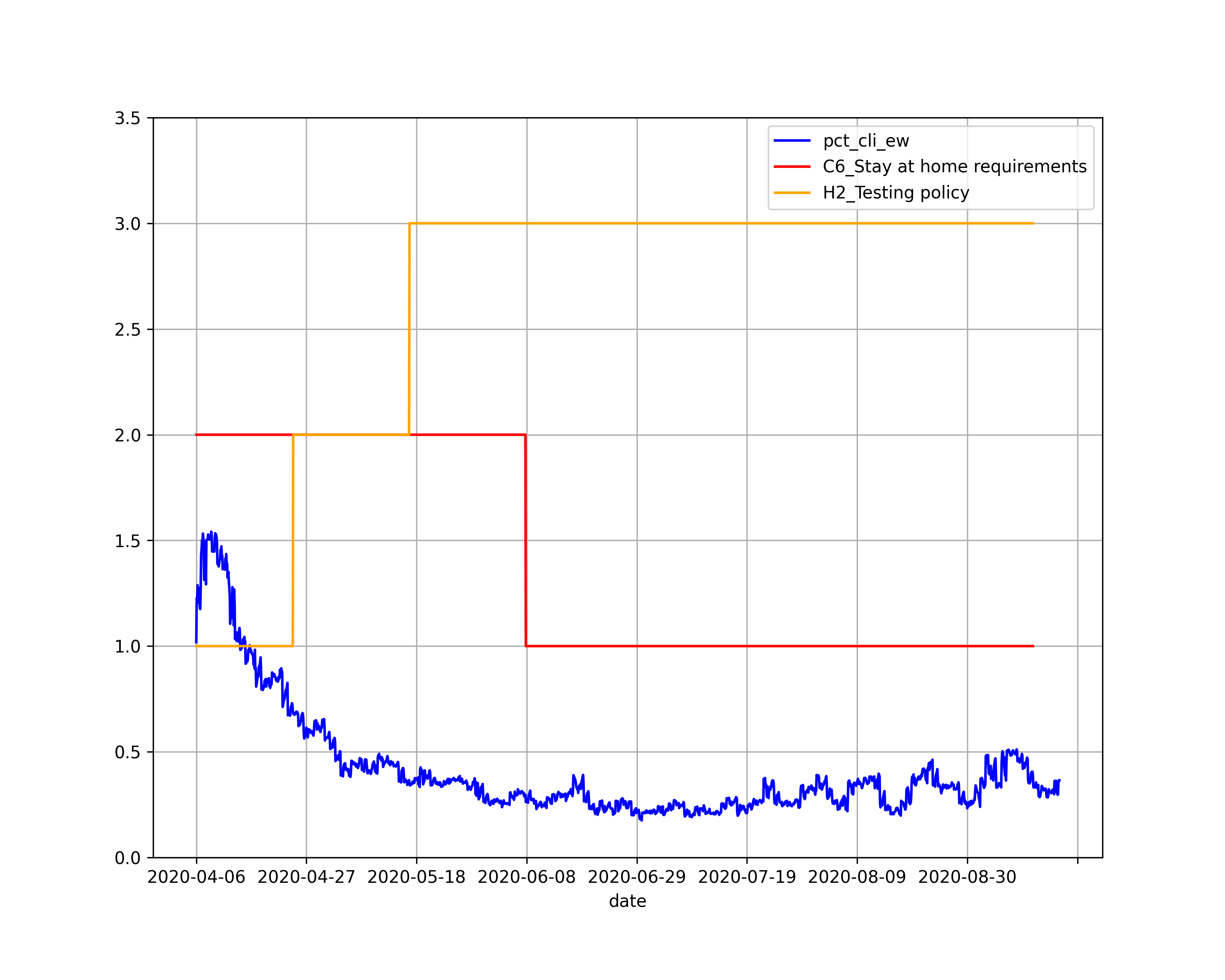}
  \caption{Policy Impacts: when Stay at home restrictions were stronger, even with higher testing rates, the \% of population with CLI (pct\_cli\_ew) was having a downward trend.}%
  \label{fig:ts_policy}
\end{figure}

\textbf{Policies vs CLI/Community Sick Impacts}: The impacts of different  non pharmaceutical interventions (NPIs) could be analysed by combining the CMU, UMD data and Oxford data \cite{OxfordDataset}. A particular analysis from that is reported here, where we notice that lifting of stay at home restrictions resulted in a sudden spike in the number of cases. This can be visualised in figure \ref{fig:ts_policy}.  \newline

\textbf{Error Metric}: We calculate 2 error metrics - \begin{itemize}
    \item Mean Absolute Error (MAE): It the absolute value of difference between predicted value and actual value, averaged over all data points. \\
     MAE = $\frac{1}{n}\sum_{i=1}^{n}|y_i - x_i|$ \\ where n is the total data instances, $y_i$ is the predicted value and $x_i$ is the actual value. 
    \item Mean Relative error (MRE): Relative Error is the absolute difference between predicted value and actual value, divided by the actual value. Mean Relative Error is Relative error averaged over all the data points.  \\
    MRE = $\frac{100}{n}\sum_{i=1}^{n}\left |\frac{y_i - x_i}{x_i +1}\right|$\\
    We add 1 in the denominator to avoid division by 0. The 100 in the numerator is to get percentage value. 
\end{itemize}

We find that a low MAE value is misleading in the case of predicting the spread of the virus; the MAE for outbreak prediction is low and has a small range (1-1.4) but more than 75\% of the target lies between 0-2.6, meaning only a small percentage of the entire population has COVID-19 (if 1\% of the entire population is affected then and MAE of 1 indicates the predicted cases could be double of actual cases). Hence, MRE is a better metric to judge a system as it accounts for even minute changes (errors) in the prediction.  \newline

\begin{figure}[t!]
    \includegraphics[width = 0.7\linewidth]{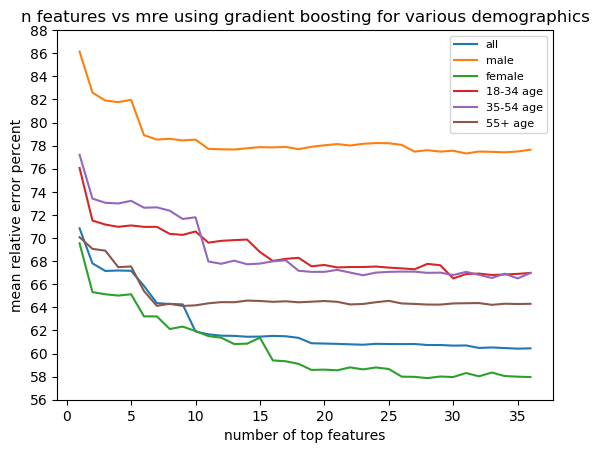}
  \caption{Error vs number of top features used for the gradient boosting model. Errors vary across demographics, and generally decrease with increase in "n". The decrease is not considerable after n = 20. }%
  \label{fig:combined}
\end{figure}

\textbf{Outbreak prediction on CMU Dataset}: Gradient boosting performs the best and considerably better than the next best algorithm in terms of the error metrics for every demographic group. Hence, only the results for Gradient Boosting are shown. Table \ref{table:symptoms} shows best accuracy achieved per dataset. For every dataset, the best "n" is in 30s. We achieve an MRE of 60.40\% for the entire dataset. The performance is better on Female-only data when compared to Male-only data. The performance is slightly better on 55+ age data than other age groups. This can also be observed from figure \ref{fig:combined}.   \newline

\begin{figure}[t!]
    \includegraphics[width =1\linewidth]{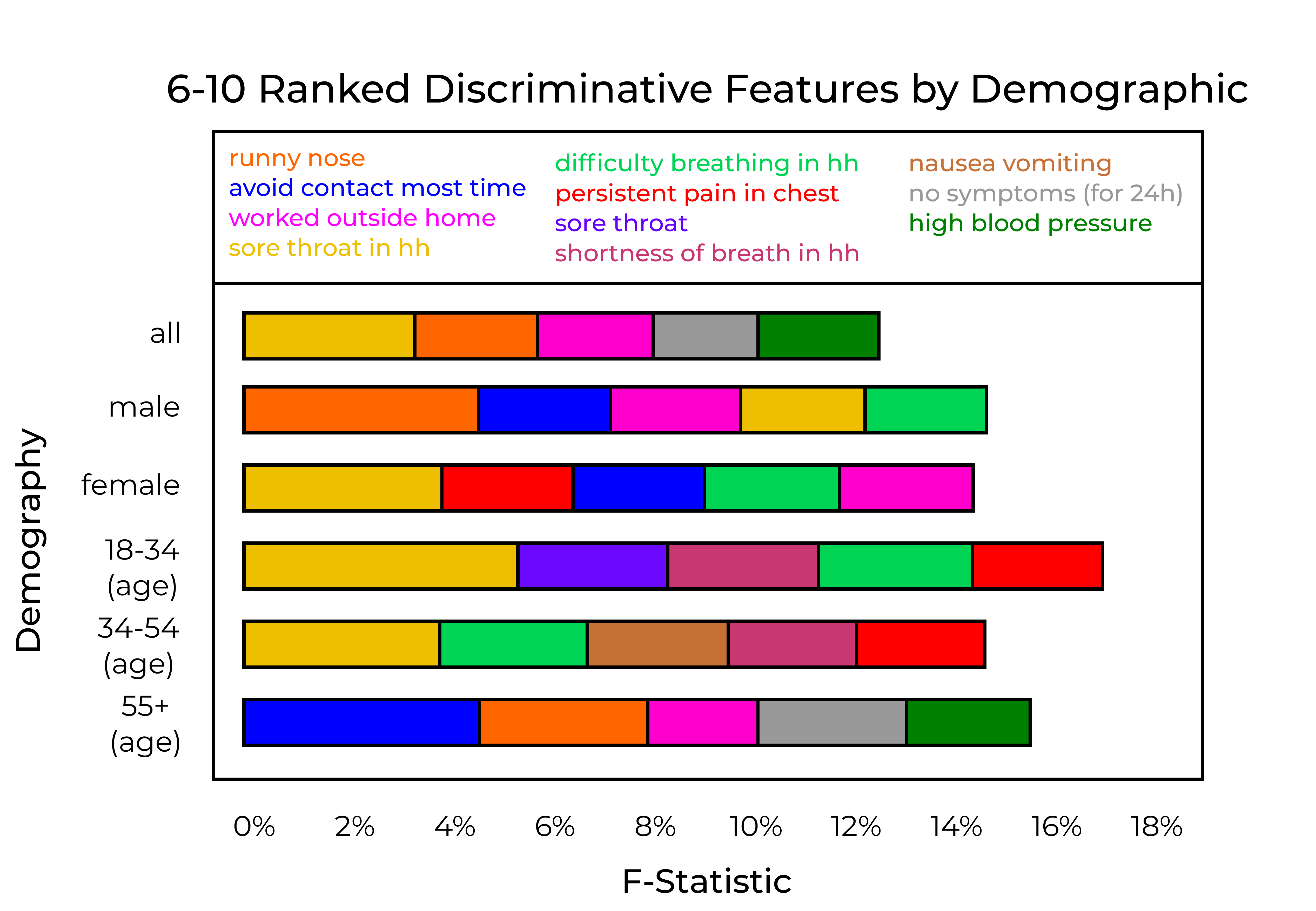}
  \caption{After the top 5 predictive features (which are roughly identical), there are considerable differences between the most predictive features segmented across demographics. For example for the age 34-55 demographic, 'sore throat in hh (household)' is the 6th most predictive feature but it is not there even in the top 10 most predictive features for the 55+ age demographic.  }%
  \label{fig:features}
\end{figure}

\textbf{Top Features}: Except for minor reordering, the top 5 features are - CLI in community, loss of smell, CLI in house hold (HH), fever in HH, fever across every data split. Top 6-10 features per data split are given in table \ref{fig:features}. We can see that 'worked outside home' and 'avoid contact most time' are useful features for male, female and 55+ age data. Figure \ref{fig:combined} shows MRE vs number of features selected for different data splits. Overall, the error decreases as we add more features. However, the decrease in error isn't very considerable when we go beyond 20 features (	$<$ 1\%).  \newline

\textbf{Time Series Analysis}: As seen in Tables \ref{tab:my-table-time-series-cmu-var}, \ref{tab:my-table-time-series-cmu-lstm}, \ref{tab:my-table-umd-var} and \ref{tab:my-table-umd-lstm},  we are able to forecast the PCT\_CLI with an MRE of 15.11\% using just 23 features from the UMD dataset. We can see that VAR performs better than LSTM on an average. This can be explained by the dearth of data available. Furthermore, we can see that the outbreak forecasting on New York was done with 11.28\% MRE, making use of only 10 features. This might be caused by an inherent bias in the sampling strategy or participant responses. For example, the high correlation noted between anosmia and COVID-19 prevalence suggests several probable causes of confounding relationships between the two. This could also occur if both symptoms are specific and sensitive for COVID-19 infection.  \newline

\begin{table}[]
\begin{tabular}{|l|l|l|}
\hline
\multirow{2}{*}{\textbf{Location}} & \multicolumn{2}{c|}{\textbf{VAR (\%)}}                  \\ \cline{2-3} 
                                   & \textit{\textbf{MRE}}           & \textit{\textbf{MAE}} \\ \hline
New York                           & 11.28, 95\% CI {[}10.9, 11.6{]} & 0.15                  \\ \hline
California                         & 13.48, 95\% CI {[}13.4, 13.5{]} & 0.23                  \\ \hline
Florida                            & 17.49, 95\% CI {[}17.5, 17.5{]} & 0.38                  \\ \hline
New Jersey                         & 17.93, 95\% CI {[}17.9, 18{]}   & 0.26                  \\ \hline
\end{tabular}
\caption{The errors of forecasting the outbreak of COVID-19 (\% of people tested and positive) for the next 30 days using VAR model.}
\label{tab:my-table-time-series-cmu-var}
\end{table}

\begin{table}[]{%
\begin{tabular}{|l|l|l|}
\hline
\multirow{2}{*}{\textbf{Location}} & \multicolumn{2}{c|}{\textbf{LSTM (\%)}}                                                                             \\ \cline{2-3} 
                                   & \textit{\textbf{MRE}} & \textit{\textbf{MAE}} \\ \hline
New York   & 23.61, 95\% CI {[}23.6, 23.7{]} & 0.36 \\ \hline
California & 45.06, 95\% CI {[}45, 45.2{]}   & 0.91 \\ \hline
Florida    & 64.98, 95\% CI {[}64.8, 65.1{]} & 1.51 \\ \hline
New Jersey & 15.78, 95\% CI {[}15.7, 15.9{]} & 0.26 \\ \hline
\end{tabular}%
}
\caption{The errors of forecasting the outbreak of COVID-19 (\% of people tested and positive) for the next 30 days using LSTM model.}
\label{tab:my-table-time-series-cmu-lstm}
\end{table}


\begin{table}[]
\begin{tabular}{|l|l|l|}
\hline
\multirow{2}{*}{\textbf{Location}} & \multicolumn{2}{c|}{\textbf{VAR (\%)}}                  \\ \cline{2-3} 
                                   & \textit{\textbf{MRE}}           & \textit{\textbf{MAE}} \\ \hline
Tokyo                              & 17.77, 95\% CI {[}17.7, 17.8{]} & 0.28                  \\ \hline
British Columbia                   & 21.35, 95\% CI {[}21.3, 21.4{]} & 0.34                  \\ \hline
Northern Ireland                   & 42.72, 95\% CI {[}42.7, 42.8{]} & 0.87                  \\ \hline
Lombardia                          & 15.31, 95\% CI {[}15.3, 15.4{]} & 0.22                  \\ \hline
\end{tabular}
\caption{Results of forecasting the outbreak of COVID-19 (\% of people with COVID-19 like illness in the population - PCT\_CLI) for the next 30 days using VAR model}
\label{tab:my-table-umd-var}
\end{table}

\begin{table}[]
\begin{tabular}{|l|l|l|}
\hline
\multirow{2}{*}{\textbf{Location}} & \multicolumn{2}{c|}{\textbf{LSTM (\%)}}                                                 \\ \cline{2-3} 
                                   & \multicolumn{1}{c|}{\textit{\textbf{MRE}}} & \multicolumn{1}{c|}{\textit{\textbf{MAE}}} \\ \hline
Tokyo            & 30.00, 95\% CI {[}29.9, 30.1{]} & 0.53 \\ \hline
British Columbia & 31.11, 95\% CI {[}30.9, 31.3{]} & 0.56 \\ \hline
Northern Ireland & 42.46, 95\% CI {[}42.1, 42.9{]} & 1.21 \\ \hline
Lombardia        & 16.11, 95\% CI {[}16, 16.2{]}   & 0.21 \\ \hline
\end{tabular}
\caption{Results of forecasting the outbreak of COVID-19 (\% of people with COVID-19 like illness in the population - PCT\_CLI) for the next 30 days using LSTM model}
\label{tab:my-table-umd-lstm}
\end{table}

\textbf{Symptoms vs CLI overlap} : The percentage of population with symptoms like cough, fever and runny nose is much higher than the percentage of people who suffer from CLI or the percentage of people who are sick in the community. Only 4\% of the people in the UMD dataset who reported to have CLI weren't suffering from chest pain and nausea.   \newline

\textbf{Ablation Studies} : Here, we perform ablation studies to verify and investigate the relative importance of the features that were selected using f regression feature ranking algorithm \cite{Freg}. In the following experiments the top $N=10$ features obtained from the f\_regression algorithm are considered as the subset for evaluation.  \newline

\begin{figure}[t!]
    \includegraphics[width=\linewidth]{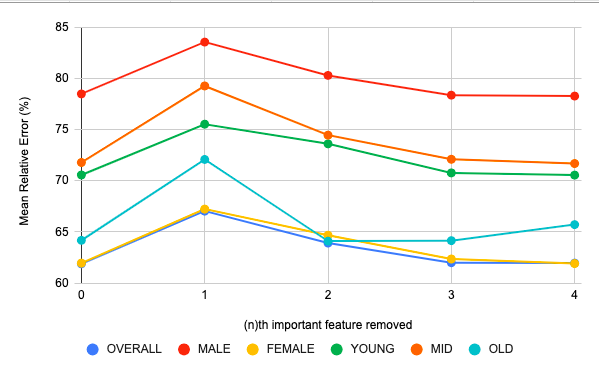}
    \caption{Results of All-but-One experiment (MRE)}
    \label{fig:mre_vs_ifr}
\end{figure}

\textbf{All-but-One}:
In this experiment, the target variable which is the percentage of people affected by COVID 19 is estimated by considering $N-1$ features from a given set of top $N$ features by dropping 1 feature at a time in every iteration in a descending order. The results are visualised in figure \ref{fig:mre_vs_ifr} from which it is clear that there is a considerable increase error when the most significant feature is dropped and the loss in performance is not as drastic when any other feature is dropped. This reaffirms our feature selection method.  \newline

\begin{figure}[t!]
    \centering
    \includegraphics[width=\linewidth]{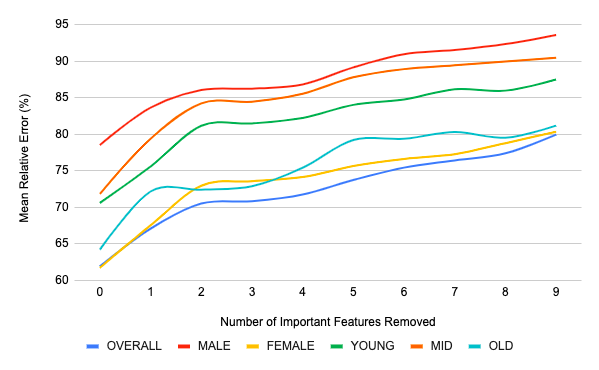}
    \caption{Results of Cumulative Feature Dropping}
    \label{fig:re_fr}
\end{figure}

\textbf{Cumulative Feature Dropping}: 
In this experiment, we estimate the target variable based on top $N$=10 features and then carry out the experiment with $N-i$ features in every iteration where $i$ is the iteration count. The features are dropped in the descending order. Figure \ref{fig:re_fr} shows the results. The change in slope from the start to the end of the graph strongly supports our previous inference that the most important feature has a huge significance on the performance and error rate and reaffirms our features selection algorithm.

\section{Conclusion And Future Work}
In this work, we analyse the benefits of COVID-19 self reported symptoms present in the  CMU, UMD, and Oxford datasets. We present correlation analysis, outbreak prediction, and time series prediction of the percentage of respondents with positive COVID-19 tests and the percentage of respondents who show COVID-like illness. By clustering datasets across different demographics, we reveal micro and macro level insights into the relationship between symptoms and outbreaks of COVID-19. These insights might form the basis for future analysis of the epidemiology and manifestations of COVID-19 in different patient populations. Our correlation and prediction studies identify a small subset of features that can predict measures of COVID-19 prevalence to a high degree of accuracy. Using this, more efficient surveys can be designed to measure only the most relevant features to predict COVID-19 outbreaks. Shorter surveys will increase the likelihood of respondent participation and decrease the chances that respondents providing false (or incorrect) information. We believe that our analysis will be valuable in shaping health policy and in COVID-19 outbreak predictions for areas with low levels of testing by providing prediction models that rely on self-reported symptom data. As shown from our results, the predictions from our models could be reliably used by health officials and policymakers, in order to prioritise resources. Furthermore, having crowdsourced information as the base, it helps to scale this method at a much higher pace, if and when required in the future (due to the advent of a newer virus or a strain).  \newline

In the future, we plan to use advanced deep learning models for predictions. Furthermore, given the promise shown by population level symptoms data we find more relevant and timely problems that can be solved with individual data. Building machine learning systems on data from mobile/wearable devices can be built to understand users' vitals, sleep behavior etc., have the data shared at an individual level, can augment the participatory surveillance dataset and thereby the predictions made. This can be achieved without compromising on the privacy of the individual.  We also plan to compare the reliability of such survey methods with actual number of cases in the corresponding regions and it's generalisability across the population.

\section{Acknowledgement}

We acknowledge the inputs of Seojin Jang, Chirag Samal, Nilay Shrivastava, Shrikant Kanaparti, Darshan Gandhi and Priyanshi Katiyar. We further thank Prof. Manuel Morales (University of Montreal), Morteza Asgari and Hellen Vasques for helping in developing a dashboard to showcase the results. We also acknowledge Dr. Thomas C. Kingsley (Mayo Clinic) for his suggestions in the future works.

\bibliography{references}

\begin{thebibliography}{25}
\providecommand{\natexlab}[1]{#1}
\providecommand{\url}[1]{\texttt{#1}}
\providecommand{\urlprefix}{URL }
\expandafter\ifx\csname urlstyle\endcsname\relax
  \providecommand{\doi}[1]{doi:\discretionary{}{}{}#1}\else
  \providecommand{\doi}{doi:\discretionary{}{}{}\begingroup
  \urlstyle{rm}\Url}\fi

\bibitem[{Fre(2007-2020)}]{Freg}
 2007-2020.
\newblock \emph{sklearn f regression}.
\newblock
  \url{https://scikit-learn.org/stable/modules/generated/sklearn.feature_selection.f_regression.html}.

\bibitem[{CDC(2020)}]{CDC}
CDC. 2020.
\newblock \emph{Data and Statistics}.
\newblock \url{https://www.cdc.gov/obesity/data/prevalence-maps.html}.

\bibitem[{Chan et~al.(2020)}]{Chan2020}
Chan, C.~C.; et~al. 2020.
\newblock Type I interferon sensing unlocks dormant adipocyte inflammatory
  potential.
\newblock \emph{Nature Communications} 11(1).
\newblock ISSN 2041-1723.
\newblock \urlprefix\url{https://doi.org/10.1038/s41467-020-16571-4}.

\bibitem[{Cheung and Lai(1995)}]{cheung1995lag}
Cheung, Y.-W.; and Lai, K.~S. 1995.
\newblock Lag order and critical values of the augmented Dickey--Fuller test.
\newblock \emph{Journal of Business \& Economic Statistics} 13(3): 277--280.

\bibitem[{Delphi~group(2020)}]{CMUDataset}
Delphi~group, C. M.~U. 2020.
\newblock Delphi's COVID-19 Surveys.
\newblock \urlprefix\url{https://covidcast.cmu.edu/surveys.html}.

\bibitem[{Fan et~al.(2020)}]{UMDDataset}
Fan, J.; et~al. 2020.
\newblock COVID-19 World Symptom Survey Data API.

\bibitem[{Friedman(2001)}]{gdbt}
Friedman, J.~H. 2001.
\newblock {Greedy function approximation: A gradient boostingmachine.}
\newblock \emph{The Annals of Statistics} 29(5): 1189 -- 1232.
\newblock \doi{10.1214/aos/1013203451}.
\newblock \urlprefix\url{https://doi.org/10.1214/aos/1013203451}.

\bibitem[{Galton(1886)}]{linear_regression}
Galton, F. 1886.
\newblock Regression Towards Mediocrity in Hereditary Stature.
\newblock \emph{The Journal of the Anthropological Institute of Great Britain
  and Ireland} 15: 246--263.
\newblock ISSN 09595295.
\newblock \urlprefix\url{http://www.jstor.org/stable/2841583}.

\bibitem[{Garcia-Agundez et~al.(2021)Garcia-Agundez, Ojo, Hern{\'a}ndez-Roig,
  Baquero, Frey, Georgiou, Goessens, Lillo, Menezes, Nicolaou
  et~al.}]{garcia2021estimating}
Garcia-Agundez, A.; Ojo, O.; Hern{\'a}ndez-Roig, H.~A.; Baquero, C.; Frey, D.;
  Georgiou, C.; Goessens, M.; Lillo, R.~E.; Menezes, R.; Nicolaou, N.; et~al.
  2021.
\newblock Estimating the COVID-19 Prevalence in Spain with Indirect Reporting
  via Open Surveys.
\newblock \emph{Frontiers in Public Health} 9.

\bibitem[{Gers, Schmidhuber, and Cummins(1999)}]{gers1999learning}
Gers, F.~A.; Schmidhuber, J.; and Cummins, F. 1999.
\newblock Learning to forget: Continual prediction with LSTM.
\newblock \emph{1999 Ninth International Conference on Artificial Neural
  Networks ICANN 99.} .

\bibitem[{Gostic et~al.(2020)Gostic, Gomez, Mummah, Kucharski, and
  Lloyd-Smith}]{Gostic}
Gostic, K.; Gomez, A.~C.; Mummah, R.~O.; Kucharski, A.~J.; and Lloyd-Smith,
  J.~O. 2020.
\newblock Estimated effectiveness of symptom and risk screening to prevent the
  spread of COVID-19.
\newblock \emph{eLife} 9.
\newblock ISSN 2050-084X.
\newblock \doi{10.7554/elife.55570}.
\newblock \urlprefix\url{https://europepmc.org/articles/PMC7060038}.

\bibitem[{Hale et~al.(2020)Hale, Webster, Petherick, Phillips, and
  Kira}]{OxfordDataset}
Hale, T.; Webster, S.; Petherick, A.; Phillips, T.; and Kira, B. 2020.
\newblock Oxford COVID-19 Government Response Tracker Blavatnik School of
  Government.

\bibitem[{Holden(1995)}]{holden1995vector}
Holden, K. 1995.
\newblock Vector auto regression modeling and forecasting.
\newblock \emph{Journal of Forecasting} 14(3): 159--166.

\bibitem[{Liang et~al.(2020)Liang, Liang, Ou, Chen, Chen, Li, Li, Guan, Sang,
  Lu, Xu, Chen, Guo, Guo, Chen, Zhao, Li, Zhang, Zhong, He, and for the China
  Medical Treatment Expert Group~for COVID-19}]{Liang}
Liang, W.; Liang, H.; Ou, L.; Chen, B.; Chen, A.; Li, C.; Li, Y.; Guan, W.;
  Sang, L.; Lu, J.; Xu, Y.; Chen, G.; Guo, H.; Guo, J.; Chen, Z.; Zhao, Y.; Li,
  S.; Zhang, N.; Zhong, N.; He, J.; and for the China Medical Treatment Expert
  Group~for COVID-19. 2020.
\newblock {Development and Validation of a Clinical Risk Score to Predict the
  Occurrence of Critical Illness in Hospitalized Patients With COVID-19}.
\newblock \emph{JAMA Internal Medicine} 180(8): 1081--1089.
\newblock ISSN 2168-6106.
\newblock \doi{10.1001/jamainternmed.2020.2033}.
\newblock \urlprefix\url{https://doi.org/10.1001/jamainternmed.2020.2033}.

\bibitem[{Menni et~al.(2020{\natexlab{a}})Menni, Valdes, Freidin, Sudre,
  Nguyen, Drew, Ganesh, Varsavsky, Cardoso, El-Sayed~Moustafa, Visconti, Hysi,
  Bowyer, Mangino, Falchi, Wolf, Ourselin, Chan, Steves, and Spector}]{Menni}
Menni, C.; Valdes, A.~M.; Freidin, M.~B.; Sudre, C.~H.; Nguyen, L.~H.; Drew,
  D.~A.; Ganesh, S.; Varsavsky, T.; Cardoso, M.~J.; El-Sayed~Moustafa, J.~S.;
  Visconti, A.; Hysi, P.; Bowyer, R. C.~E.; Mangino, M.; Falchi, M.; Wolf, J.;
  Ourselin, S.; Chan, A.~T.; Steves, C.~J.; and Spector, T.~D.
  2020{\natexlab{a}}.
\newblock Real-time tracking of self-reported symptoms to predict potential
  COVID-19.
\newblock \emph{Nature Medicine} 26(7): 1037--1040.
\newblock ISSN 1546-170X.
\newblock \doi{10.1038/s41591-020-0916-2}.
\newblock \urlprefix\url{https://doi.org/10.1038/s41591-020-0916-2}.

\bibitem[{Menni et~al.(2020{\natexlab{b}})}]{menni2020real}
Menni, C.; et~al. 2020{\natexlab{b}}.
\newblock Real-time tracking of self-reported symptoms to predict potential
  COVID-19.
\newblock \emph{Nature medicine} 1--4.

\bibitem[{NIH(2020)}]{BMI}
NIH. 2020.
\newblock \emph{Adult Body Mass Index (BMI)}.
\newblock
  \url{https://www.nhlbi.nih.gov/health/educational/lose_wt/BMI/bmicalc.htm}.

\bibitem[{Parma et~al.(2020)}]{Parma}
Parma, V.; et~al. 2020.
\newblock More than smell. COVID-19 is associated with severe impairment of
  smell, taste, and chemesthesis.
\newblock \emph{medRxiv} \doi{10.1101/2020.05.04.20090902}.
\newblock
  \urlprefix\url{https://www.medrxiv.org/content/early/2020/05/24/2020.05.04.20090902}.

\bibitem[{Pedregosa et~al.(2011)}]{scikit-learn}
Pedregosa, F.; et~al. 2011.
\newblock Scikit-learn: Machine Learning in {P}ython.
\newblock \emph{Journal of Machine Learning Research} 12: 2825--2830.

\bibitem[{Quinlan(1986)}]{dtrees}
Quinlan, J.~R. 1986.
\newblock Induction of decision trees.
\newblock \emph{Machine Learning} 1(1): 81--106.
\newblock ISSN 1573-0565.
\newblock \doi{10.1007/BF00116251}.
\newblock \urlprefix\url{https://doi.org/10.1007/BF00116251}.

\bibitem[{Rodriguez et~al.(2020{\natexlab{a}})Rodriguez, Muralidhar, Adhikari,
  Tabassum, Ramakrishnan, and Prakash}]{rodriguez2020steering}
Rodriguez, A.; Muralidhar, N.; Adhikari, B.; Tabassum, A.; Ramakrishnan, N.;
  and Prakash, B.~A. 2020{\natexlab{a}}.
\newblock Steering a Historical Disease Forecasting Model Under a Pandemic:
  Case of Flu and COVID-19.
\newblock \emph{arXiv preprint arXiv:2009.11407} .

\bibitem[{Rodriguez et~al.(2020{\natexlab{b}})Rodriguez, Tabassum, Cui, Xie,
  Ho, Agarwal, Adhikari, and Prakash}]{Rodriguezdeep}
Rodriguez, A.; Tabassum, A.; Cui, J.; Xie, J.; Ho, J.; Agarwal, P.; Adhikari,
  B.; and Prakash, B.~A. 2020{\natexlab{b}}.
\newblock DeepCOVID: An Operational Deep Learning-driven Framework for
  Explainable Real-time COVID-19 Forecasting.
\newblock \emph{medRxiv} \doi{10.1101/2020.09.28.20203109}.
\newblock
  \urlprefix\url{https://www.medrxiv.org/content/early/2020/09/29/2020.09.28.20203109}.

\bibitem[{Saad-Roy et~al.(2020)}]{saad2020immune}
Saad-Roy, C.~M.; et~al. 2020.
\newblock Immune life history, vaccination, and the dynamics of SARS-CoV-2 over
  the next 5 years.
\newblock \emph{Science} .

\bibitem[{Shi et~al.(2020)}]{Shi}
Shi, F.; et~al. 2020.
\newblock Review of Artificial Intelligence Techniques in Imaging Data
  Acquisition, Segmentation and Diagnosis for COVID-19.
\newblock \emph{IEEE Reviews in Biomedical Engineering} 1–1.
\newblock ISSN 1941-1189.
\newblock \doi{10.1109/rbme.2020.2987975}.
\newblock \urlprefix\url{http://dx.doi.org/10.1109/RBME.2020.2987975}.

\bibitem[{Wilder, Mina, and Tambe(2020)}]{wilder2020tracking}
Wilder, B.; Mina, M.~J.; and Tambe, M. 2020.
\newblock Tracking disease outbreaks from sparse data with Bayesian inference.
\newblock \emph{arXiv preprint arXiv:2009.05863} .

\end{thebibliography}

%
\section{Appendix}
\onecolumn

The sample features present in the datasets can be observed in table \ref{data-sample-signals}. 

\begin{table}[t]
\begin{tabular}{|l|l|}
\hline
\textbf{Dataset}                     & \textbf{Example Signals}                                                \\
\hline
UMD                                  & COVID-like illness symptoms, influenza-like illness symptoms, mask usage    \\
\hline
CMU                                  & sore throat, loss of smell/taste, chronic lung disease                      \\
\hline
OxCGRT                               & containment and closure policies, economic policies, health system policies \\
\hline
\end{tabular}
\caption{Example Signal Information for the Datasets}
\label{data-sample-signals}
\end{table}

\subsection{Correlation Studies}
\label{sec:corr_appens}
\begin{figure}[h!]
\centering
\graphicspath{ {./images/} }
\begin{center}
\includegraphics[scale=0.38]{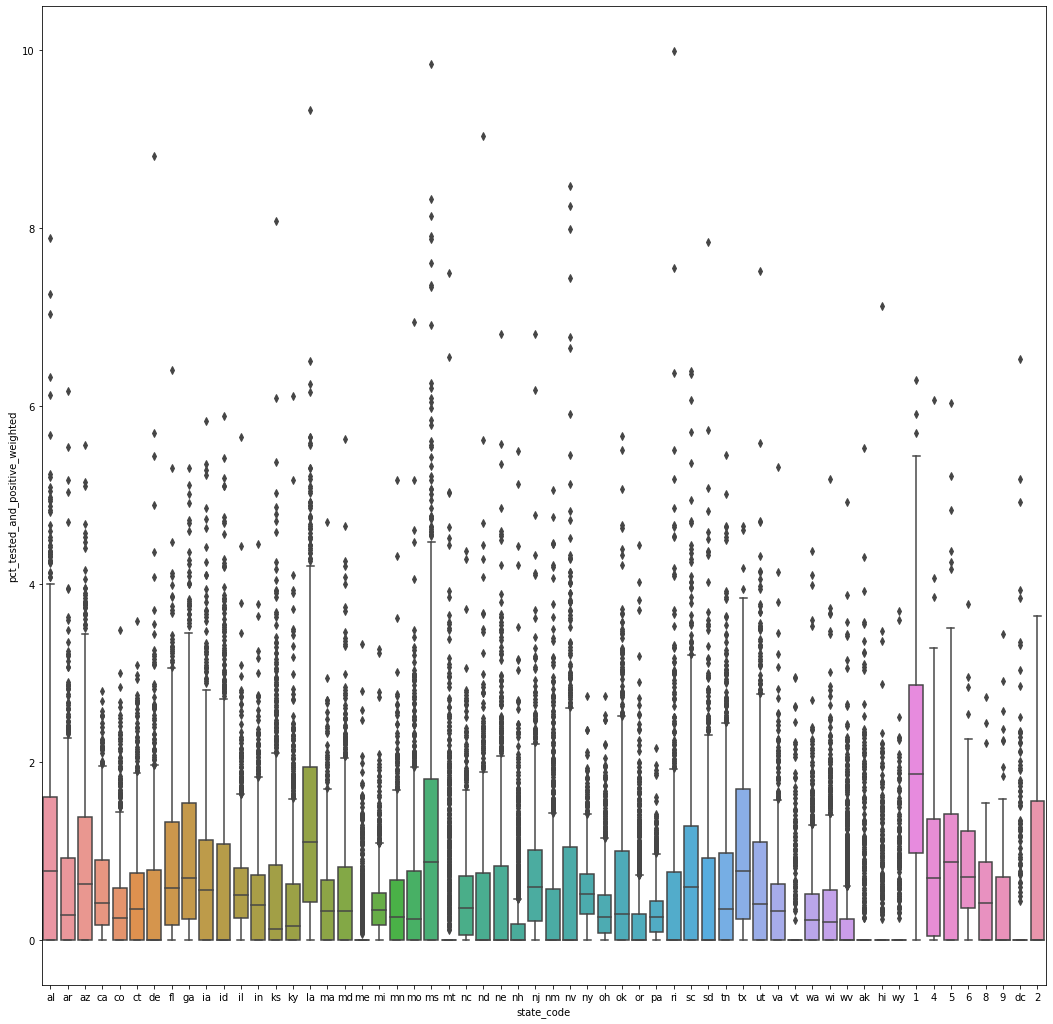}
\end{center}
\caption{State wise distribution of percentage of people tested COVID positive}
\label{fig:cmucorr2}
\end{figure}

The detailed plots of the correlation analysis of the CMU dataset is noted in figure \ref{fig:cmucorr2}.

\begin{figure}[ht]
\centering
\graphicspath{ {./images/} }
\begin{center}
\includegraphics[scale=0.4]{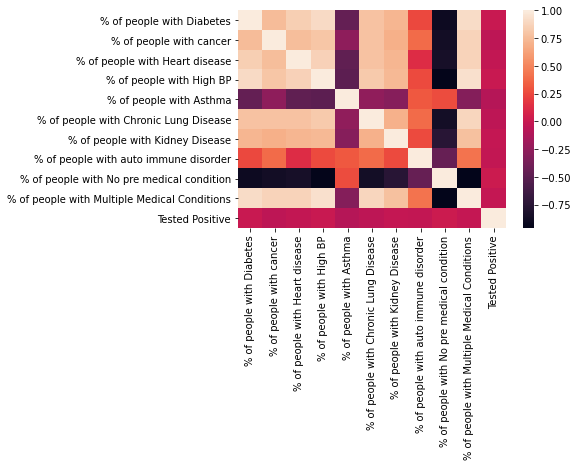}
\end{center}
\caption{Correlation study: The relationship between the underlying medical condition  and percentage People tested COVID positive}
\label{fig:cmucorr_medical}
\end{figure}


\begin{figure}[h!]
\centering
\graphicspath{ {./images/} }
\begin{center}
\includegraphics[scale=0.22]{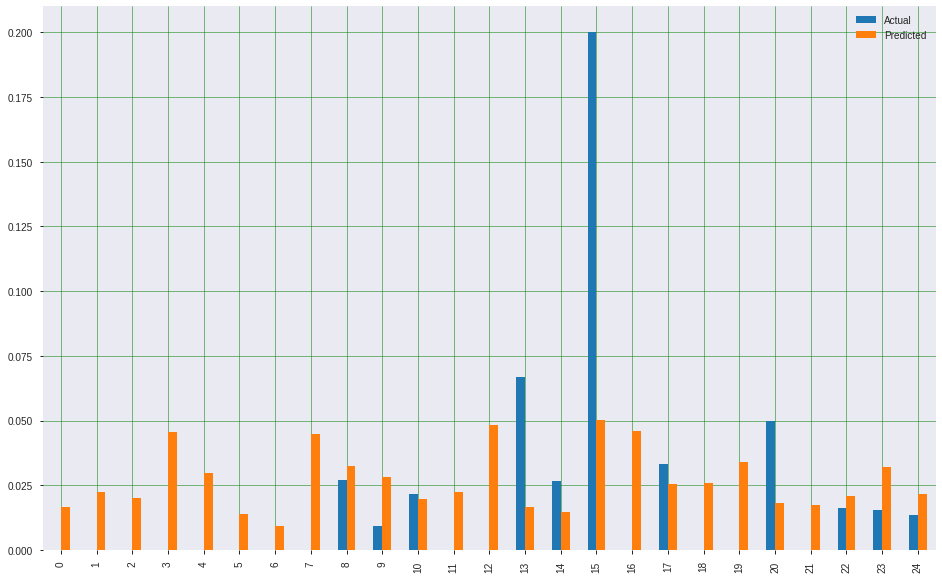}
\end{center}
\caption{Observed VS Predicted: Prediction of percentage of people tested positive using percentage of people who recently had a contact with someone who is COVID positive.  }
\label{fig:cmucorr_contact}
\end{figure}

\begin{figure}[h!]
\centering
\graphicspath{ {./images/} }
\begin{center}
\includegraphics[width=15cm,height=60cm,keepaspectratio]{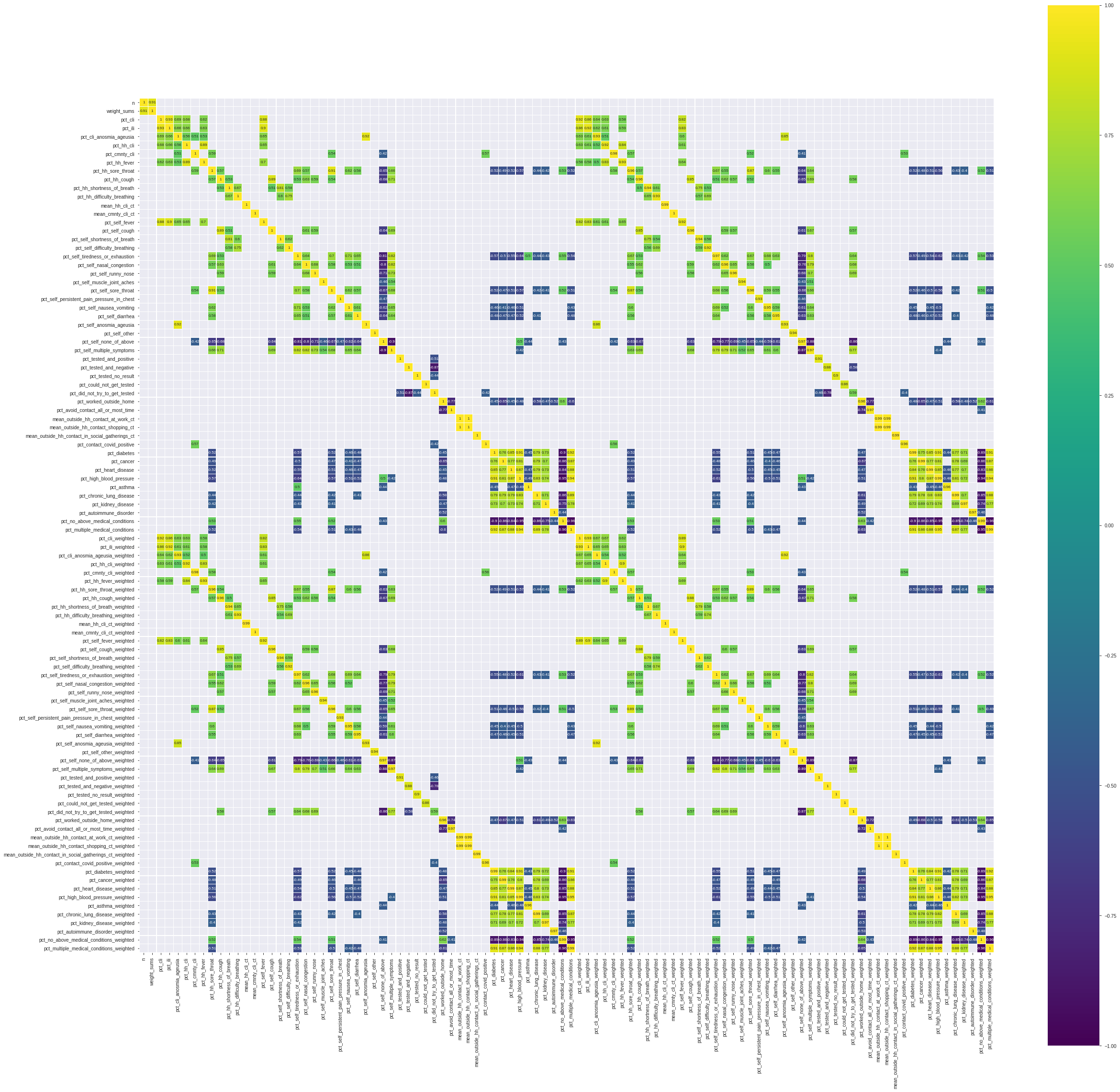}
\end{center}
\caption{Correlation map depicting the relationship between the features along with the target variable(s)}
\label{fig:cmucorr_target}
\end{figure}

\begin{figure}[h!]
\centering
\graphicspath{ {./images/} }
\begin{center}
\includegraphics[scale=0.4]{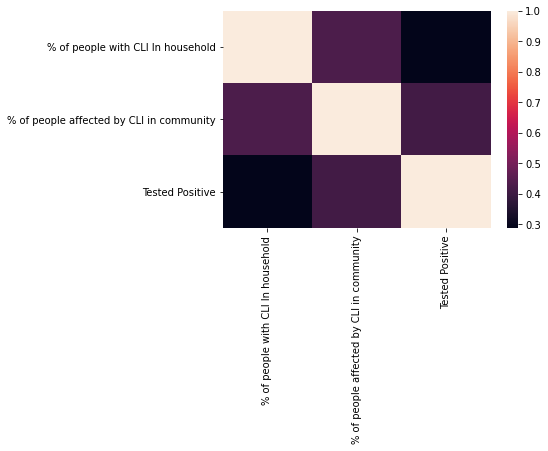}
\end{center}
\caption{Correlation study: The relationship between the COVID like illness and percentage People tested COVID positive}
\label{fig:cmucorr_cli}
\end{figure}

\label{appendix}

\begin{table}[ht]
\centering
\begin{tabular}{l|l|l}
\textbf{Rank} & \textbf{Signal}                                            & \textbf{F\_Statistic} \\
\hline
1             & COVID-like Illness in Community                                  & 14938.48816456    \\
2             & Loss of smell or taste                      & 9498.89229794     \\
3             & COVID-like Illness in Household                                & 6050.88250153     \\
4             & Fever in Household                                   & 5490.15612527     \\
5             & Fever                                 & 4388.95759983     \\
6             & Sore Throat in Household                            & 1787.42269067     \\
7             & Avoid contact with others most of the time         & 1494.25038393     \\
8             & Difficulty breathing in Household                   & 1330.48793481     \\
9             & Persistent Pain Pressure in Chest & 1257.78331468     \\
10            & Runny Nose                           & 1084.84412662     \\
11            & Worked outside home                       & 1023.50285601     \\
12            & Nausea or Vomiting                      & 1016.94758914     \\
13            & Shortness of breath in Household                   & 1004.67944587     \\
14            & Sore Throat                          & 975.25614266      \\
15            & Difficulty Breathing                 & 723.49150048      \\
16            & Asthma                                      & 466.91243179      \\
17            & Shortness of Breath                 & 440.88344033      \\
18            & Cough in Household                                   & 322.05679444      \\
19            & No symptoms in past 24 hours                       & 241.72819985      \\
20            & Diarrhea                              & 228.59465358      \\
21            & Chronic Lung Disease                     & 224.24651285      \\
22            & Cancer                                      & 205.19827073      \\
23            & Other Pre-existing Disease                                 & 158.31567587      \\
24            & Tiredness or Exhaustion             & 134.36715409      \\
25            & Cough                                 & 84.66549815       \\
26            & No Above Medical Conditions              & 84.40193799       \\
27            & Heart Disease                              & 74.71994609       \\
28            & Multiple Medical Conditions               & 52.61630823       \\
29            & Autoimmune Disorder                        & 40.8942176        \\
30            & Nasal Congestion                     & 33.60170138       \\
31            & Kidney Disease                             & 23.88450351       \\
32            & Average people in Household with COVID-like ilness                                & 14.52969291       \\
33            & Multiple Symptoms                    & 12.56805547       \\
34            & Muscle Joint Aches                  & 1.72398411        \\
35            & High Blood Pressure                       & 0.48328156        \\
36            & Diabetes                                    & 0.24390025       
\end{tabular}
\caption{Features ranked for entire by \textit{F score}. All signals are represented as percentages of respondents who responded that way.}
\label{tab:ranked-features}
\centering
\end{table}

\subsection{Time Series}

In table \ref{table:examples-umd} we continue to experiment with different spatial windows, like trying to predict PCT\_CLI for different locations like "Tokyo" and "British Columbia" using different combination of features. Further on table \ref{table:examples-us} analysis is done on more US states with an LSTM based deep learning model to predict PCT\_CLI and we notice that there is no significant gain in using DL models (probably due to lack of data).
The pct\_community\_sick is another variable which we try to predict, and the results can be seen in table \ref{table:examples-cmnty}

\begin{figure}[h!]
\centering
\graphicspath{ {./pca_plots/} }
\begin{center}
\includegraphics[scale=0.4]{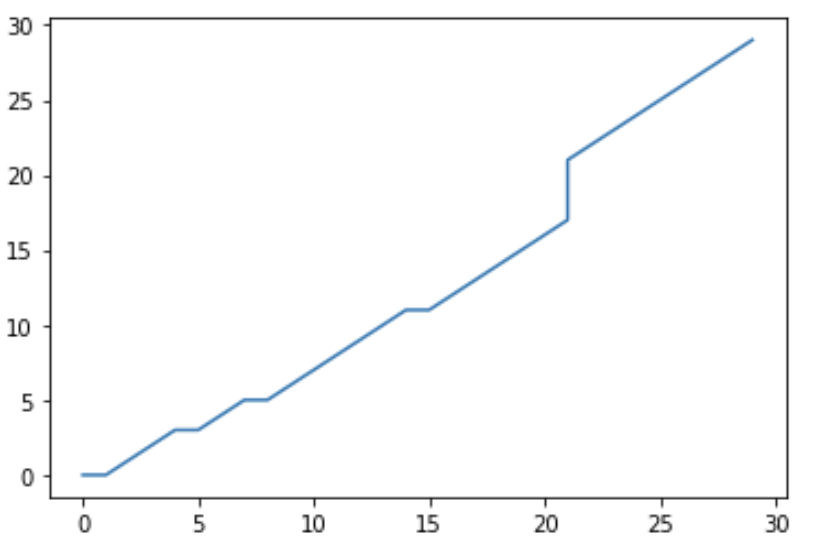}
\end{center}
\caption{DTW plot analysing the relationship between our forecasted curve vs the original curve for Ohio State}
\label{fig:dtw-oh}
\end{figure}

\begin{figure}[h!]
\centering
\graphicspath{ {./pca_plots/} }
\begin{center}
\includegraphics[scale=0.4]{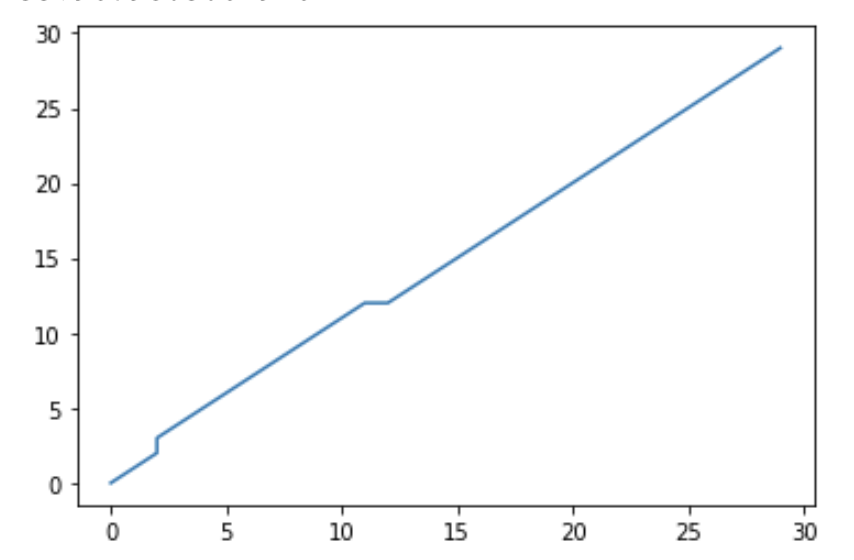}
\end{center}
\caption{DTW plot analysing the relationship between our forecasted curve vs the original curve}
\label{fig:dtw-tx}
\end{figure}

In figs [\ref{fig:dtw-oh},\ref{fig:dtw-tx}] we do Dynamic Time Warping(DTW) to compare how well our forecasted timeseries curve matches with the original curve. DTW was used due to the flexibility to compare timeseries signals which are of different lengths. This will enable us to compare different temporal windows across different spatial windows to understand the effectiveness of the model at different contexts.
\begin{table}[ht]
\centering
\begin{tabular}{|p{2cm}|p{2cm}|p{1cm}|p{1cm}|p{1cm}|p{3cm}|}
 \hline
 \textbf{Location} & \textbf{Bucket} 
 & \textbf{RMSE} & \textbf{MAE} & \textbf{MRE (\%)} & \textbf{Features Used}\\
 \hline
 \textit{Abu Dhabi} & male and age 18-34   & 2.43 & 2.23 & 167.86 &difficulty breathing + anosmia ageusia (weighted)\\
 \hline
\textit{Tokyo} & female and age 35-54 & 0.56 &  0.47 & 30.16 & difficulty breathing + anosmia ageusia (weighted)\\
 \hline
 \textit{British Columbia} & male and age 55+ &  1.09 & 0.59 & 28.68 &difficulty breathing + anosmia ageusia (weighted)\\
 \hline
 \textit{Lombardia} & male and age 55+ & 0.95  & 0.67 & 28.72 & difficulty breathing + anosmia ageusia (weighted) \\
 \hline
 \textit{Lombardia} & male and age 55+ & 0.95  & 0.67 & 28.72 & Behavioural / external features (weighted) \\
 \hline
 \textit{British Columbia} & male and age 55+ &  1.07 & 0.76 & 50.17 & Behavioural / external features (weighted)\\
 \hline
 \textit{Tokyo} & female and age 35-54 & 0.58 &  0.49 & 31.38 & Behavioural / external features (weighted)\\
 \hline
  \textit{Abu Dhabi} & male and age 18-34   & 2.91 & 2.78 & 207.94 &Behavioural / external features (weighted)\\
 \hline
\end{tabular}
\caption{RMSE and MAE scores for different buckets of interest + Ablation - VAR model - PCT \_CLI \_weighted }
\label{table:examples-umd}
\end{table}

\begin{table}[ht]
\centering
\begin{tabular}{|p{3cm}|p{3cm}|p{1cm}|p{1cm}|p{1cm}|}
 \hline
 \textbf{Location} & \textbf{Bucket} 
 & \textbf{RMSE} & \textbf{MAE} & \textbf{MRE (\%)}\\
 \hline
 \textit{Abu Dhabi} & male and age 18-34   & 9.99 & 8.94 & 73.11 \\
 \hline
\textit{Tokyo} & female and age 35-54 & 1.13 & 1.02  & 41.67 \\
 \hline
 \textit{British Columbia} & male and age 55+ & 3.21 & 2.65 & 137.13\\
 \hline
 \textit{Lombardia} & male and age 55+ & 1.25 & 1.25  & 24.49 \\
 \hline
\end{tabular}
\caption{RMSE and MAE scores for different buckets of interest - VAR model - PCT \_Community \_Sick }
\label{table:examples-cmnty}
\end{table}

\begin{table}[ht]
\centering
\begin{tabular}{|p{3cm}|p{2cm}|p{1cm}|p{1cm}|p{1cm}|p{2cm}|}
 \hline
 \textbf{Location} & \textbf{Bucket} 
 & \textbf{RMSE} & \textbf{MAE} &\textbf{MRE} &\textbf{Model}\\
 \hline
 \textit{TX} & male and age overall & 1.56 & 1.21 & 43.00 & VAR\\
 \hline
\textit{CA} & male and age overall & 1.22  & 0.93 & 23.44 & VAR\\
 \hline
 \textit{NY} & female and age overall & 0.7 & 0.56 & 21.59 & VAR\\
 \hline
 \textit{FL} & female and age overall & 1.48 & 1.18 & 19.35 & VAR\\
 \hline
 
 \textit{TX} & male and age overall & 6.28 & 4.06 & 89.4 & LSTM\\
 \hline
\textit{CA} & male and age overall & 2.83 & 2.68 & 71.24 & LSTM\\
 \hline
 \textit{NY} & female and age overall & 2.02  & 1.9 & 68.17 & LSTM\\
 \hline
 \textit{FL} & female and age overall & 4.33 & 4.19 & 73.34 & LSTM\\
  \hline
\end{tabular}
\caption{RMSE and MAE scores for different buckets of interest - VAR/LSTM models - PCT\_CLI - Here we see that deep learning models aren't performing better than normal statistical models}
\label{table:examples-us}
\end{table}

\begin{figure}[h!]
\centering
\graphicspath{ {./pca_plots/} }
\begin{center}
\includegraphics[scale=0.4]{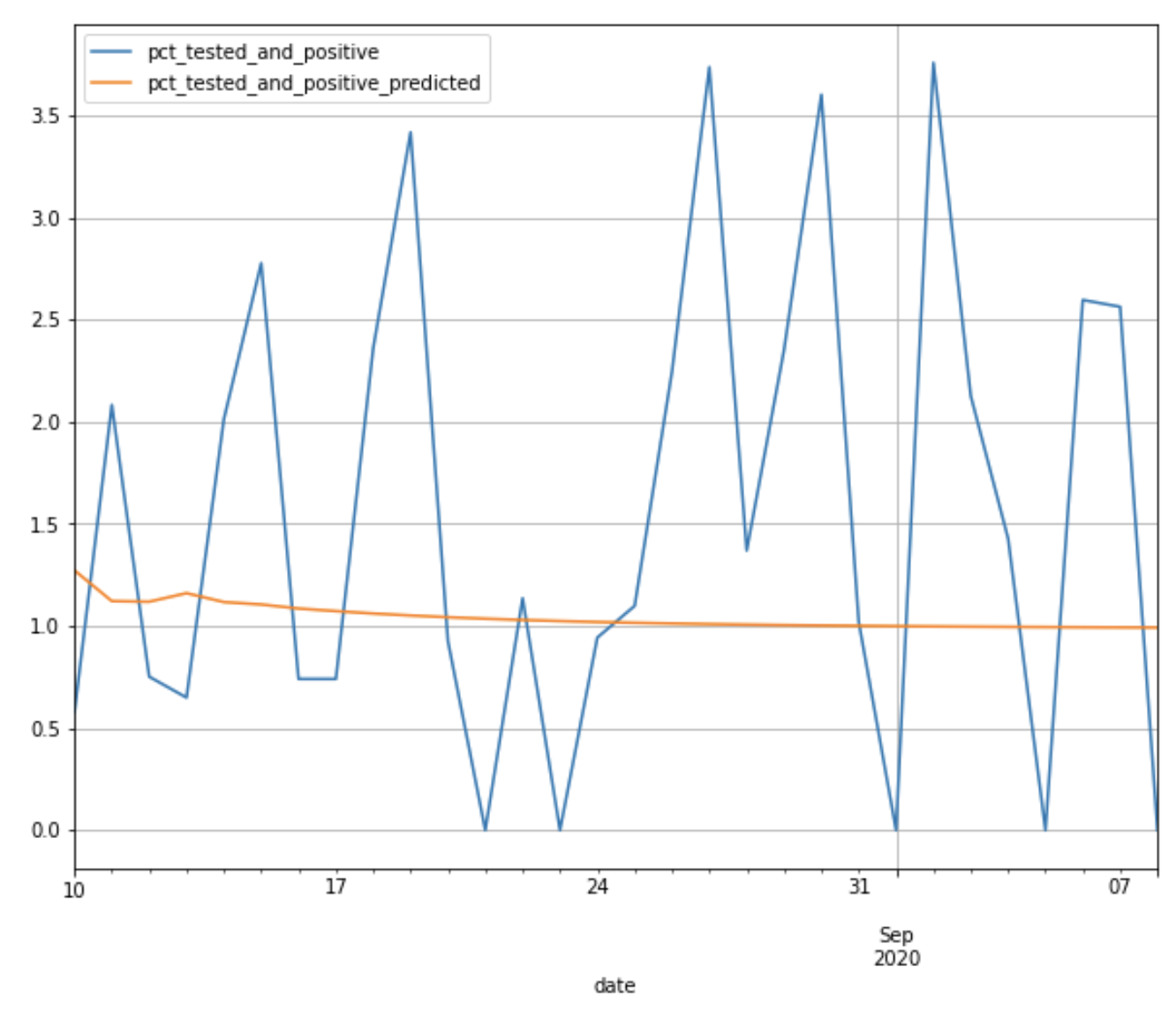}
\end{center}
\caption{Forecasted curve vs the original curve for Ohio.}
\label{fig:pred-oh}
\end{figure}

\begin{figure}[ht]
\centering
\graphicspath{ {./pca_plots/} }
\begin{center}
\includegraphics[scale=0.4]{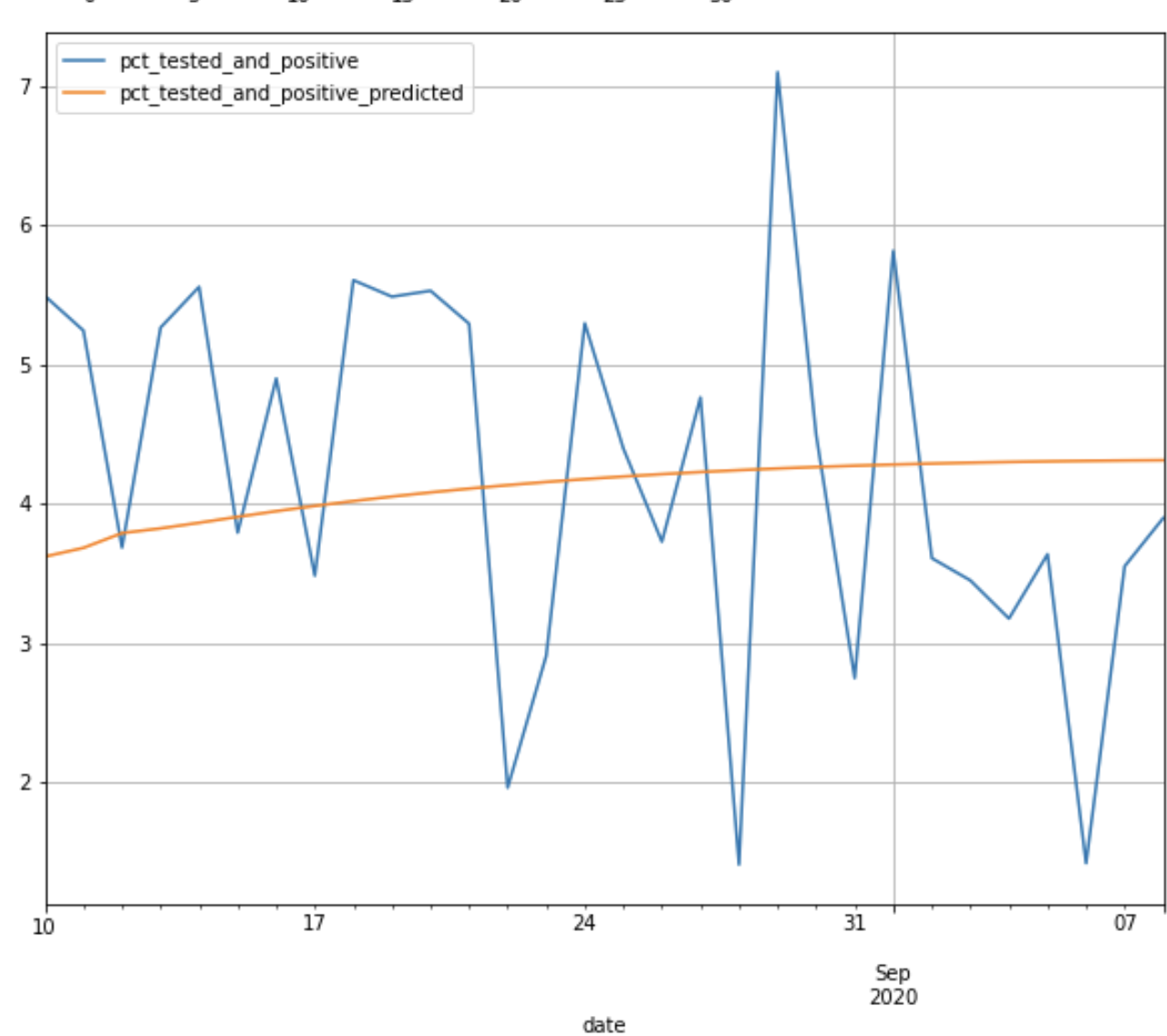}
\end{center}
\caption{Forecasted curve vs the original curve for Texas.}
\label{fig:pred-tx}
\end{figure}

\begin{table}
\centering
\begin{tabular}{|l|l|l|l|} 
\hline
\textbf{Demography} & \textbf{Feature Removed} & \textbf{MAE} & \textbf{MRE} \\ 
\hline
\hline
\multirow{5}{*}{\textbf{Male}} & no feature removed & 1.389806313 & 77.42367322 \\ 
\cline{2-4}
 & pct\_cmnty\_cli\_weighted & 1.470745054 & 82.97970974 \\ 
\cline{2-4}
 & pct\_self\_anosmia\_ageusia\_weighted & 1.423361929 & 79.90430572 \\ 
\cline{2-4}
 & pct\_self\_none\_of\_above\_weighted & 1.410196471 & 78.62630177 \\ 
\cline{2-4}
 & pct\_self\_runny\_nose\_weighted & 1.398427829 & 78.13485192 \\ 
\hline
\hline
\multirow{5}{*}{\textbf{Female}} & no feature removed & 1.100879926 & 57.63336087 \\ 
\cline{2-4}
 & pct\_cmnty\_cli\_weighted & 1.218554308 & 64.54253671 \\ 
\cline{2-4}
 & pct\_self\_anosmia\_ageusia\_weighted & 1.155647687 & 61.12311515 \\ 
\cline{2-4}
 & pct\_self\_none\_of\_above\_weighted & 1.121811889 & 58.73118158 \\ 
\cline{2-4}
 & pct\_self\_runny\_nose\_weighted & 1.104380112 & 57.92685018 \\ 
\hline
\hline
\multirow{5}{*}{\textbf{Young}} & no feature removed & 1.231519891 & 67.07207641 \\ 
\cline{2-4}
 & pct\_cmnty\_cli\_weighted & 1.31846811 & 72.201516 \\ 
\cline{2-4}
 & pct\_self\_anosmia\_ageusia\_weighted & 1.277138933 & 70.38556851 \\ 
\cline{2-4}
 & pct\_avoid\_contact\_all\_or\_most\_time\_weighted & 1.244334089 & 67.80402144 \\ 
\cline{2-4}
 & pct\_self\_runny\_nose\_weighted & 1.234101952 & 67.46623764 \\ 
\hline
\hline
\multirow{5}{*}{\textbf{Mid }} & no feature removed & 1.276053866 & 67.05778653 \\ 
\cline{2-4}
 & pct\_cmnty\_cli\_weighted & 1.384547554 & 73.44381028 \\ 
\cline{2-4}
 & pct\_self\_anosmia\_ageusia\_weighted & 1.326526868 & 70.22181485 \\ 
\cline{2-4}
 & pct\_self\_none\_of\_above\_weighted & 1.321293709 & 69.44829708 \\ 
\cline{2-4}
 & pct\_avoid\_contact\_all\_or\_most\_time\_weighted & 1.285893087 & 67.62940495 \\ 
\hline
\hline
\multirow{5}{*}{\textbf{Old}} & no feature removed & 1.172592164 & 63.98633923 \\ 
\cline{2-4}
 & pct\_cmnty\_cli\_weighted & 1.314221647 & 72.59134309 \\ 
\cline{2-4}
 & pct\_avoid\_contact\_all\_or\_most\_time\_weighted & 1.191250701 & 64.98442049 \\ 
\cline{2-4}
 & pct\_self\_anosmia\_ageusia\_weighted & 1.192677984 & 65.76644281 \\ 
\cline{2-4}
 & pct\_self\_multiple\_symptoms\_weighted & 1.186357275 & 64.7244507 \\
\hline
\end{tabular}
\end{table}

\begin{table}
\centering
\begin{tabular}{|l|l|l|l|c} 
\hline
\textbf{Demography} & \textbf{Feature Removed} & \textbf{MAE} & \textbf{MRE} \\ 
\hline
\hline
\multirow{11}{*}{\textbf{Overall}} & no feature removed & 1.143995128 & 60.83421503 \\ 
\cline{2-4}
 & pct\_cmnty\_cli\_weighted & 1.248043237 & 67.08605954 \\ 
\cline{2-4}
 & pct\_self\_anosmia\_ageusia\_weighted & 1.177417511 & 63.07033879 \\ 
\cline{2-4}
 & pct\_self\_none\_of\_above\_weighted & 1.169464223 & 61.67148756 \\ 
\cline{2-4}
 & pct\_self\_runny\_nose\_weighted & 1.149200232 & 61.32185068 \\ 
\cline{2-4}
 & pct\_hh\_cli\_weighted & 1.14551667 & 60.93481883 \\ 
\cline{2-4}
 & pct\_avoid\_contact\_all\_or\_most\_time\_weighted & 1.149772918 & 61.16631628 \\ 
\cline{2-4}
 & pct\_worked\_outside\_home\_weighted & 1.147615986 & 61.0433573 \\ 
\cline{2-4}
 & pct\_self\_fever\_weighted & 1.144711565 & 60.92739832 \\ 
\cline{2-4}
 & pct\_hh\_fever\_weighted & 1.143703022 & 60.7946325 \\ 
\cline{2-4}
 & pct\_hh\_difficulty\_breathing\_weighted & 1.143007815 & 60.83204654 \\
\hline
\end{tabular}
\end{table}

\end{document}